\documentclass[letterpaper, 10 pt, conference]{ieeeconf}  

\usepackage{graphicx}
\usepackage{cite}
\usepackage{caption}
\usepackage{xcolor}
\usepackage{amsmath}
\usepackage{amssymb}
\usepackage{booktabs}
\usepackage{multirow}
\usepackage{multicol}
\usepackage{subcaption}
\usepackage{hyperref}
\hypersetup{hidelinks}
\usepackage{booktabs,multirow,tabularx,array}

\IEEEoverridecommandlockouts                              

\title{\LARGE \bf
PanoTac: A Dense Three-Axis Tactile Sensor 
\\for Robust Touch-Centric Manipulation
}

\author{
    Binglin (Andre) Wang$^{1,\dagger}$,
    Xiaole (Alex) Jiang$^{2,\dagger}$,
    Zicheng Teng$^{1}$,
    Xinqi Liu$^{1}$,\\
    Yunzhu Li$^{3}$,
    Lerrel Pinto$^{1}$,
    Zichen Jeff Cui$^{1,\ddagger}$,
    and Binghao Huang$^{3,\ddagger,*}$
}

\begin{document}

\makeatletter
\twocolumn[{%
  \begin{@twocolumnfalse}
  \@maketitle

  {\centering
  \small
  $^{1}$New York University
  $^{2}$City University of New York\\
  \quad
  $^{3}$Columbia University
  \par}
  \begin{center}
    \includegraphics[width=\textwidth]{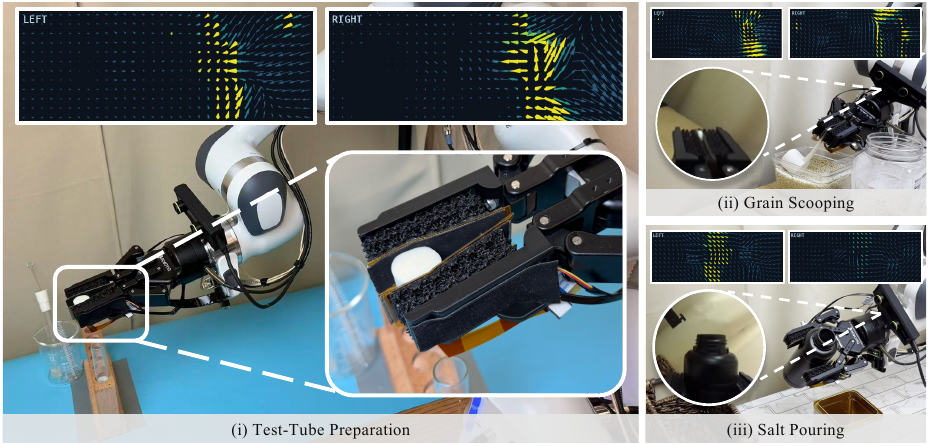}
    \captionof{figure}{We propose \textbf{PanoTac}, a low-cost, compliant tactile fingertip for dense three-axis tactile sensing. We showcase PanoTac with three tasks \textit{(i)} test-tube preparation, where in-hand state helps infer tube orientation and shear feedback helps detect stopper release; \textit{(ii)} grain scooping, where contact patterns help infer spoon orientation and guide robot--environment interaction; and \textit{(iii)} salt pouring, where weight cues distinguish filled from empty bottles and inform pouring decisions.}
    \label{fig:teaser_v4}
  \end{center}
  \vspace{-0em}
  \end{@twocolumnfalse}
}]
\begingroup
    
  \renewcommand{\thefootnote}{}
  \footnotetext[0]{%
    \begin{tabular}{@{}l@{\enspace}l@{}}
      $\dagger$  & Equal contribution. \\
      $\ddagger$ & Equal advising. \\
      $*$        & Corresponding author: Binghao Huang. \\
                 & Email: \texttt{binghao.huang@columbia.edu}.
    \end{tabular}%
  }
\endgroup
\setcounter{footnote}{0}

\makeatother
\thispagestyle{empty}
\pagestyle{empty}

\begin{abstract}

Contact-rich manipulation benefits from tactile feedback that captures both detailed contact geometry and interaction forces.
When scooping grains, contact patterns help infer spoon orientation, while resistance transmitted through the grasp guides motion.
To provide these complementary sensing capabilities, we present \textit{PanoTac}, a low-cost, compliant tactile fingertip designed for dense three-axis tactile sensing. Each fingertip provides a $12 \times 32 \times 3$ tactile representation, with one normal ($z$) and two shear ($x$, $y$) response components at each taxel. The sensor operates at 60\,Hz and costs less than \$50 per fingertip. Sensor characterization quantifies response repeatability and consistency across sensor instances and contact locations.
We integrate \textit{PanoTac} into a robotic system with a visuo-tactile pipeline. Across test-tube preparation, grain scooping, and salt pouring, the resulting policies outperform vision-only baselines by an average of 46.7 percentage points. Ablations show that removing either spatial contact information or shear feedback reduces overall performance across the task suite. Our project page is available at \href{https://panotac.github.io}{https://panotac.github.io}.

\end{abstract}

\section{Introduction}
\label{sec:intro}



Humans rely on vision and touch to manipulate daily objects, using tactile feedback to understand contact states and determine how an interaction should proceed. Consider inserting a test tube into a rack and removing its stopper (Fig.~\ref{fig:teaser_v4}(i)). Vision guides positioning, while contact patterns help infer the tube's in-hand orientation under partial occlusion. Stopper removal requires sensing whether the stopper remains engaged or has come free. Changes in resistance transmitted through the fingers help us decide when to stop wiggling and lift the stopper free. These stages highlight two complementary aspects of touch: spatial contact information helps estimate in-hand object orientation, while shear feedback helps detect changes in interaction state. 


These requirements motivate dense three-axis tactile sensing, but realizing this capability in a practical robotic fingertip presents two challenges. \textit{(i) Spatial resolution and force sensing.} Dense piezoresistive arrays provide detailed responses to normal contact but do not directly capture shear loading~\cite{3d-vitac,flexitac, touchinthewild}. Magnetic tactile sensors provide responses to normal and shear loading~\cite{eflesh,anyskin}, but the size and spacing of discrete sensing elements constrain spatial sampling density, making fine contact patterns difficult to resolve directly~\cite{hou2025supermag}. \textit{(ii) Hardware integration and reproducibility.} Optical tactile sensors capture detailed contact geometry and support force estimation, but accommodating the camera and optical path imposes constraints on fingertip packaging and geometry~\cite{gelsight}. Practical deployment also requires affordable, compliant hardware with straightforward fabrication~\cite{sipos2024gelslim}, repeatable responses, and consistency across independent instances.

To address these challenges, we present \textbf{PanoTac}, a low-cost, compliant tactile fingertip designed for dense three-axis tactile sensing. Our contributions are threefold:

\textit{(i)~A dense three-axis tactile fingertip.}
We combine spatial contact information from a dense piezoresistive sensing array~\cite{flexitac} with magnetic responses to fingertip deformation~\cite{eflesh} on a shared grid.
This yields a $12 \times 32 \times 3$ tactile representation per fingertip, with one normal ($z$) and two shear ($x$, $y$) response components at each taxel.
We design the magnet arrangement to reduce magnetic cross-talk between opposing fingertips.
The fingertip is modular and straightforward to fabricate, with each sensor costing below \$50, excluding readout boards.
Fabrication requires less than 3 hours of unattended printing and 30 minutes of hands-on assembly.

\textit{(ii)~Sensor characterization and system integration.}
We evaluate response repeatability under repeated loading and characterize consistency across sensor instances and contact locations.
We integrate \textit{PanoTac} into a robotic system that conveys tactile information through auditory feedback, enabling operators to adjust actions during data collection.

\textit{(iii)~Visuo-tactile policy learning and modality analysis.}
We learn visuo-tactile manipulation policies and evaluate them on test-tube preparation, grain scooping, and salt pouring.
The resulting policies achieve an overall success rate of 91.7\% across these three tasks, compared with 45.0\% for vision-only policies.
Through modality ablations, we evaluate the complementary contributions of dense spatial contact information and shear feedback, showing that removing either reduces overall task success.

We describe the sensor design, fabrication, tactile representation, and policy learning pipeline in Sec.~\ref{sec:system}, report hardware characterization in Sec.~\ref{sec:characterization}, and present robot learning experiments and modality ablations in Sec.~\ref{sec:learning}.

\begin{figure*}[t]
\centering
\includegraphics[width=\textwidth]{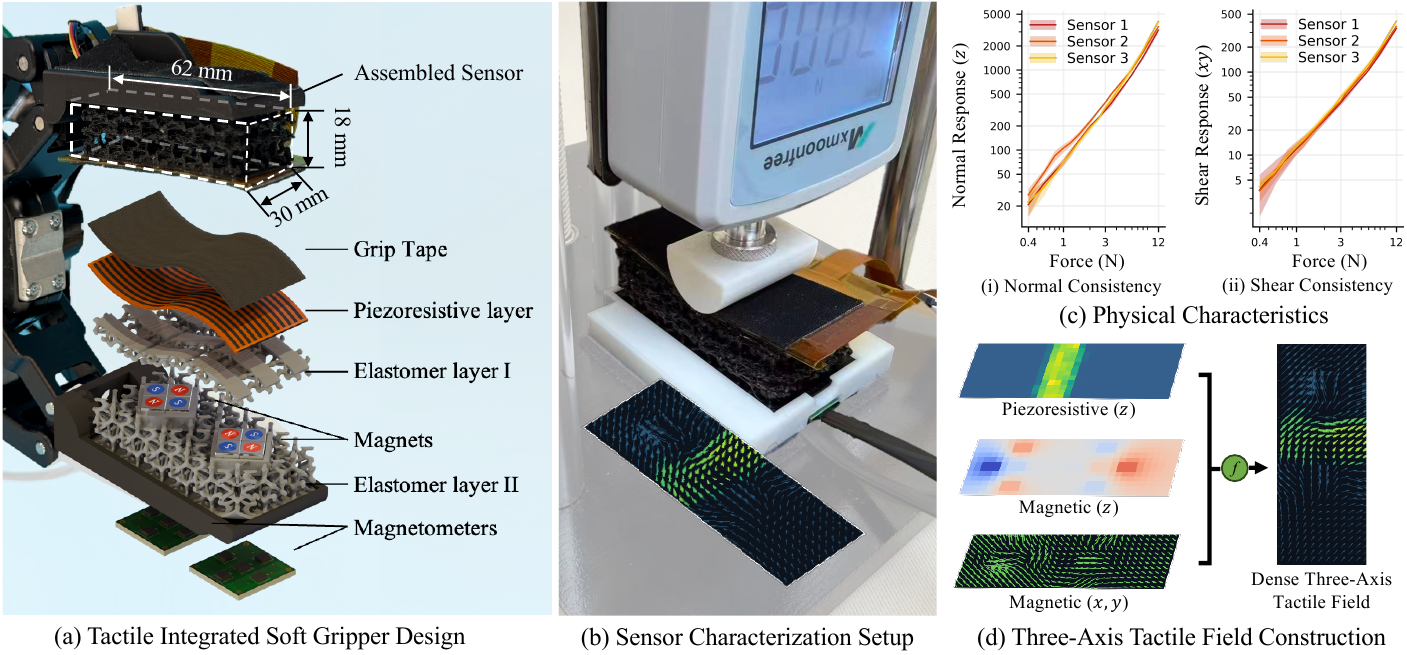}
\captionof{figure}{\textbf{PanoTac Tactile Sensing Platform.}
\textit{Part (a)} shows the assembled sensor and an exploded view of the multilayer design.
\textit{Part (b)} shows the indentation setup for sensor characterization.
\textit{Part (c)} shows the consistency of (i) the normal response $z$ and (ii) the RMS shear $(x,y)$ magnitude across three sensor under normal loads of $0.4$--$12$\,N. Shaded bands indicate $\pm 1$ standard deviation.
\textit{Part (d)} illustrates how we construct the $12 \times 32 \times 3$ dense three-axis tactile representation.}
\label{fig:hardware}
\vspace{-1em}
\end{figure*}

\section{Related Work}
\label{sec:related}
We first review tactile sensing mechanisms and the tradeoffs that motivate dense three-axis sensing. We then discuss how prior work incorporates touch into visuo-tactile manipulation and learning.

\subsection{Tactile Sensing Hardware}
\label{subsec:mechanisms}
Optical sensing is a widely used approach to tactile sensing~\cite{gelsight,digit,gelslim2018,tactip2018,omnitact} and is also incorporated into multimodal sensors~\cite{polytouch,magicgel}. These sensors have a high-dimensional observation space (an image) and extract tactile information from a camera observing deformation of the elastomer contact surface~\cite{digit360,fingereye}. While this provides detailed contact geometry, the optical path between camera and elastomer sets a minimum sensor thickness, decreasing its manipulation clearance compared to other sensing mechanisms, and the elastomer surface abrades with use and is treated as a consumable~\cite{digit}. Capacitive sensors offer a thinner profile and high sensitivity~\cite{schmitz2011skin,maiolino2013skin}, and recent designs~\cite{coinft,umift} deliver multi-axis force readings from a compact package. A practical challenge is calibration: the capacitance-to-force mapping is nonlinear and varies between units, requiring individual sensor calibration. Contact-microphone sensors~\cite{maniwav,sonicsense} detect tactile events by registering mechanical vibrations propagating through the contact surface. This yields a naturally event-driven signal, but a passive microphone carries no static component of the contact state and cannot directly measure sustained loads.

Piezoresistive arrays~\cite{stag, 3d-vitac, flexitac, osmo} are flexible, easy to fabricate, and provide a high-resolution map of normal pressure, including contact location and shape, but cannot directly measure shear. Magnetic sensors~\cite{uskin, reskin, anyskin, eflesh} infer aggregate normal and shear force from the deformation of a magnetized skin or microstructure, but their spatial resolution is bounded by the magnetometer count, so they provide a sparse local contact description. They also suffer from cross-talk between neighboring sensing units and from sensitivity to ferromagnetic objects and external fields.

Our sensor draws the best of both worlds. It keeps the low-cost fabrication of piezoresistive and microstructure-based magnetic skins, and it produces a three-axis tactile representation combining normal and shear responses with contact location and geometry. Sec.~\ref{sec:system} describes the design.

\subsection{Visuo-Tactile Manipulation Systems and Learning}
\label{subsec:multimodal}
Recent work has explored tactile representation learning and adapting generalist policies to additional sensory modalities~\cite{fuse}. One line learns a unified representation space across sensor classes~\cite{unitouch, sparsh, sparshx, t3, anytouch, tactx}, so that data collected on one sensor benefits a policy deployed on another; this is complementary to our work, and a sensor like ours is another sensor type these representations could incorporate. Another line reconstructs one tactile modality from another, to extract more tactile information from a limited sensor setup~\cite{touch2touch, li2026felt}. 
Our proposition is the opposite: when affordable hardware can directly measure the tactile information needed for the task, one should favor direct measurement over reconstruction.


Work on visuo-tactile manipulation~\cite{3d-vitac, anyskin, visk, polytouch, umift, reactive_diffusion_policy, policy_consensus, sata, foar} usually builds the learning pipeline around a particular sensor, with a fusion module or pre-trained encoder designed for it. The task suites in this line of work also tend to be vision-centric, with tactile feedback primarily refining visual information. For example, task designs often focus on one axis of tactile information, such as force feedback helping with soft-touch and insertion, or contact pose helping with reorientation. We instead treat the sensor as one more observation stream that enters a standard end-to-end policy alongside the camera. We verify it on touch-centric tasks that a vision-only policy cannot solve, such as weighing a bottle or determining the orientation of an occluded tool before using it. Further details are provided in Sec.~\ref{sec:learning}.

\section{PanoTac: Design, Fabrication, and Learning}
\label{sec:system}


\subsection{Overview}
\textbf{PanoTac} is a compliant, modular fingertip for dense three-axis tactile sensing. As shown in Fig.~\ref{fig:hardware}, its piezoresistive array captures detailed contact patterns, while magnetic measurements detect field changes caused by deformation of the supporting elastomer body. We construct a $12 \times 32 \times 3$ tactile representation per fingertip from these signals, with one normal ($z$) and two shear ($x$, $y$) response components at each taxel. This section describes the hardware design, tactile representation, and integration into a robotic system for demonstration collection and policy learning.

\subsection{Hardware Design and Layout}
\label{subsec:hardware}

As shown in Fig.~\ref{fig:hardware}(a), \textbf{PanoTac} comprises a piezoresistive sensing array, two compliant elastomer layers with magnets embedded between them, and magnetometers fixed at the base. This structure supports the contact surface while allowing deformation under normal and shear loading. The fingertip has a $62 \times 30$\,mm contact surface and an $18$\,mm-thick contact stack. Each fingertip is carried by a rigid PLA housing that attaches to a parallel-jaw gripper finger. The housing holds the magnetometer boards at the rear and locates the sensing body through a keyed front slot, allowing tool-free replacement of the assembled sensing module.

\textit{(i) Normal and shear sensing principles.}
We use the dense piezoresistive array~\cite{flexitac}, providing $12 \times 32$ taxels at a pitch of $2$\,mm. Normal loading changes the local electrical resistance measured at each taxel. To capture shear loading, we support the array with a compliant elastomer body embedded with neodymium magnets. External loading deforms the elastomer and displaces the magnets relative to the fixed magnetometers, changing the measured magnetic fields~\cite{eflesh,anyskin}.

\textit{(ii) Compliant support structure.}
To support the contact array while accommodating fingertip deformation, we use coplanar thermoplastic polyurethane (TPU) tiles separated by relief gaps. These tiles also provide bonding surfaces for the array. We use TPU~95A above the magnet pockets to support the contact array and retain the magnets.
Below the pockets, TPU~85A provides a more compliant layer that permits load-induced magnet displacement, including under shear. We apply grip tape (3M Gripping Material TB641) to the contact-facing surface to increase friction and use double-sided adhesive to bond the array to the elastomer support.

\textit{(iii) Magnet layout and interference reduction.}
We arrange two magnet groups along the fingertip, each centered within an elastomer pocket directly above its magnetometer board. This placement strengthens local signals and increases separation between neighboring groups. Each group contains four magnets in a $2 \times 2$ checkerboard pattern, with adjacent magnets magnetized in opposite directions along the surface normal to reduce magnetic interference. We rotate each group $45^\circ$ about the surface normal relative to the magnetometer axes to distribute shear-induced magnetic responses across $x$ and $y$ channels and reduce the risk of channel saturation.

To explain the far-field behavior, we model the four magnets in each group as equal-magnitude point dipoles located at the corners of a square, with magnetic moments normal to the square plane.
In the ideal symmetric checkerboard configuration, the dipole and quadrupole moments cancel, leaving an octupole as the leading multipole. For the same magnet positions and moment magnitudes, an idealized multipole expansion gives the far-field magnitudes for aligned and checkerboard configurations:
\begin{equation}
    |\mathbf{B}_{\mathrm{aligned}}|
    = B_0 \, g_{\mathrm{aligned}}(\Omega)
    \left(\frac{a}{r}\right)^3
    \left[1+\mathcal{O}\left(\frac{a^2}{r^2}\right)\right],
    \label{eq:aligned_field}
\end{equation}
\begin{equation}
    |\mathbf{B}_{\mathrm{checker}}|
    = B_0 \, g_{\mathrm{checker}}(\Omega)
    \left(\frac{a}{r}\right)^5
    \left[1+\mathcal{O}\left(\frac{a^2}{r^2}\right)\right].
    \label{eq:checker_field}
\end{equation}
Here, $2a$ is the side length of a single $2\times2$ magnet arrangement, $r$ is the distance from the center of that arrangement, and $B_0$ sets the magnetic field strength scale. The dimensionless functions $g_{\mathrm{aligned}}(\Omega)$ and $g_{\mathrm{checker}}(\Omega)$ describe the angular dependence on the observation direction $\Omega$ from that center. Under this idealized model, the leading checkerboard far-field term scales as $r^{-5}$, compared with $r^{-3}$ for the aligned configuration.
This faster decay motivates the alternating-polarity layout for reducing magnetic cross-talk between neighboring sensing units and opposing fingertips.

\subsection{Three-Axis Tactile Representation}
\label{subsec:signal}

We construct a dense three-axis tactile representation by mapping contact and magnetic measurements onto a shared taxel grid, as shown in Fig.~\ref{fig:hardware}(d). We first low-pass filter both sensing streams and subtract their respective no-contact baselines.
We denote the resulting magnetic response at magnetometer $m$ by $\Delta \mathbf{B}_m$.

\textit{(i) Normal component.}
The normal component combines spatial contact patterns from the piezoresistive array with a global magnitude from the magnetic $z$ responses. We suppress piezoresistive readings below a noise threshold and apply Gaussian smoothing with standard deviation $0.5$ taxel to obtain the contact activation map $A$. We set $N=0$ when no contact is detected. Otherwise, we normalize $A$ by its maximum and scale it by the mean absolute magnetic $z$ response across all magnetometers:
\begin{equation}
    N_{ij}
    =
    \left(
    \frac{1}{M}\sum_{m=1}^{M}
    |\Delta B_{m,z}|
    \right)
    \frac{A_{ij}}{\max_{u,v} A_{uv}},
    \label{eq:normal_map}
\end{equation}
where $M=10$ is the number of magnetometers per fingertip and $(i,j)$ indexes the taxel grid.

\textit{(ii) Shear components.}
We assign each magnetometer's $\Delta B_{m,x}$, $\Delta B_{m,y}$ readings to two taxels in the central
column of its corresponding footprint on the contact grid. We fill the remaining taxels by interpolating each component from the magnetometer centers using inverse-distance weighting with power $p=2$, obtaining shear-related responses $S^x_{ij}$ and $S^y_{ij}$.

\textit{(iii) Three-axis tactile field.}
Combining normal and shear response maps yields the three-axis tactile representation:
\begin{equation}
    \mathbf{T}_{ij} = \left(S^x_{ij}, S^y_{ij}, N_{ij}\right),
    \qquad
    \mathbf{T} \in \mathbb{R}^{12 \times 32 \times 3}.
\end{equation}
We use this field for sensor characterization and real-time tactile visualization. Throughout the paper, we use $z$ to denote the normal response component and $(x,y)$ to denote the two shear response components.  Our policy takes this three-axis representation as tactile input, as detailed in Sec.~\ref{subsec:robot_integration}.

\subsection{System Integration and Policy Learning}
\label{subsec:robot_integration}

We equip both fingers of a Robotiq~2F-85 gripper, mounted on a Franka Panda arm, with \textbf{PanoTac}.
We acquire tactile measurements at 60\,Hz.

\textit{(i) Teleoperation and data collection.}
We collect demonstrations through VR-based teleoperation. We display the contact map as a real-time heatmap and provide stereo tones whose pitch and volume increase with the deviation of each fingertip's magnetic readings from their moving average. This visual and auditory feedback helps the operator adjust robot motion during teleoperation. For each demonstration, we record wrist-camera RGB images, robot proprioception, tactile measurements, and action commands.
We align the recorded streams at 30\,Hz and process the tactile measurements as described in Sec.~\ref{subsec:signal}. Details of the teleoperation interface are given in Appendix~\ref{app:teleop}.




\textit{(ii) Policy architecture.}
We train a U-Net diffusion policy~\cite{diffusionpolicy, touchinthewild} jointly with its observation encoders to predict an action sequence from a single multimodal observation frame. 
An ImageNet-pretrained ResNet-18 encodes the visual observations and is fine-tuned jointly with the policy.
A single three-layer CNN followed by fully connected layers jointly encodes the dense three-axis tactile representations from both fingertips, constructed as described in Sec.~\ref{subsec:signal}.
The visual and tactile features are concatenated with the robot proprioception to condition the U-Net through FiLM.
For the tactile baselines (Sec.~\ref{ssec:learning_setup}), we use an MLP to encode the sparse three-axis magnetic readings and a CNN to encode measurements from the dense normal array. Encoder, training, and deployment details are given in Appendix~\ref{app:policy}.

\section{Sensor Characterization}
\label{sec:characterization}

We evaluate the repeatability and cross-instance consistency of
\textit{PanoTac} under loading
(Sec.~\ref{ssec:repeatability}) and its response variation
across contact locations (Sec.~\ref{ssec:spatial}).




\subsection{Experimental Setup and Response Metrics}
\label{ssec:setup}

We characterize \textit{PanoTac} using a force gauge with an experiment-specific indenter. For load-response experiments, the curved surface of a D-shaped indenter produces a rectangular contact area, as shown in Fig.~\ref{fig:hardware}(b). For spatial characterization, a spherical indenter produces localized contacts at selected positions on the fingertip. The force gauge provides the reference normal load for both experiments.

We record the piezoresistive and magnetic streams synchronously during each indentation and process them as described in Sec.~\ref{subsec:signal}. From these we obtain a normal response and a shear response, the latter defined as the RMS of the 2-D $(x,y)$ shear vector over all taxels. Sensor responses are derived from the readout boards' digital outputs, while applied loads are measured in newtons. Appendix~\ref{app:hardware} details the response computation and indentation protocols.

\subsection{Response Consistency under Loading}
\label{ssec:repeatability}

We test three independently fabricated sensor instances under normal loads at 15 levels from 0.4 to 12\,N, with three repeated sweeps per instance. Fig.~\ref{fig:hardware}(c) shows the normal $z$ and RMS shear $(x,y)$ responses. For each instance, lines and shaded bands indicate the mean and $\pm 1$ standard deviation across the three sweeps.

We quantify response consistency using the coefficient of variation,
$\mathrm{CV}=100\,\sigma/\mu$. At each load, within-instance CV
measures variation across the three sweeps of an instance, while
cross-instance CV measures variation across the mean responses
of the three instances. We average these CVs within three load
bands: 0.4--1\,N, 1.2--4\,N, and 6--12\,N. In Table~\ref{tab:force_consistency}, repeatability improves with load, with mean within-instance CVs reaching
$4.8\%$ for the normal $z$ response and $4.3\%$ for the shear
$(x,y)$ response at 6--12\,N. Cross-instance mean CVs for the shear
$(x,y)$ response remain below $9\%$, while those for the normal
$z$ response decrease from $22.9\%$ to $12.5\%$ and $10.8\%$
across the three load bands.

\subsection{Spatial Response Consistency}

\label{ssec:spatial}

\begin{figure}
  \centering
  \includegraphics[width=0.4\textwidth]{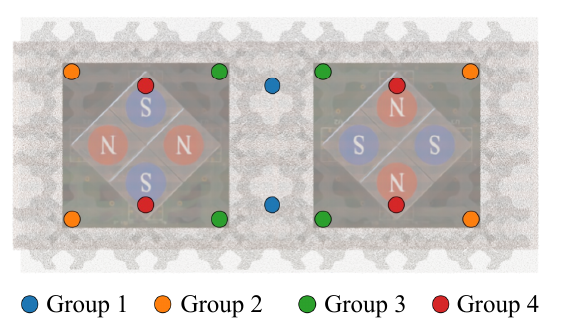}
  \caption{\textbf{Probe locations for spatial response characterization.}
Colored dots indicate probe locations, grouped according to their
positions relative to the magnet layout.
Each dot is probed 10 times.
The corresponding response statistics are reported in
Table~\ref{tab:spatial_response}.}
  \label{fig:spatial}
\end{figure}

The piezoresistive layer samples on a highly symmetric 4\,mm$^2$ taxel grid, while the magnetometers sense load only at sparse key points. Due to the nature of our magnetic sensor, the magnetometer reading enjoys fewer spatial symmetries than its piezoresistive counterpart. To probe for spatial consistency, we therefore define four probe location classes by their symmetric position on the sensor (Fig.~\ref{fig:spatial}): two points on the median line (G1), four points at the corners (G2), four points near the median line on the edge of the sensor (G3), and four interior points on the magnet groups (G4). Every site within a class is equivalent to the others under the symmetry of the magnet layout and is therefore amenable to statistical averaging. At every site we sample 10 presses. Both modalities are recorded from the same press. Table~\ref{tab:spatial_response} reports the summary statistics. The responses exhibit moderate variation relative to their
means, with response magnitudes varying across probe groups.

\begin{table}[!t]
\centering

\caption{\textbf{Force-response consistency}
($\mathrm{CV}=100\sigma/\mu$, sample SD).
Repeatability: CV over the three sweeps of one instance;
cross-instance: CV over the three instance means.
Entries average per-load CVs within each band, repeatability also over
the three instances.}

\label{tab:force_consistency}
\small
\setlength{\tabcolsep}{4pt}
\renewcommand{\arraystretch}{0.9}   
\setlength{\aboverulesep}{0.15ex}   
\setlength{\belowrulesep}{0.3ex}    
\begin{tabular}{@{}llrrr@{}}
\toprule
 & & 0.4--1\,N & 1.2--4\,N & 6--12\,N \\
\midrule
\multirow{2}{*}{Repeatability}
  & Normal & 15.5\,\% & 6.9\,\% & 4.8\,\% \\
  & Shear  & 21.1\,\% & 8.3\,\% & 4.3\,\% \\
\addlinespace[2pt]
\multirow{2}{*}{Cross-instance}
  & Normal & 22.9\,\% & 12.5\,\% & 10.8\,\% \\
  & Shear  & 4.2\,\%  & 5.3\,\% & 8.8\,\% \\
\bottomrule
\end{tabular}
\end{table}


\begin{table}[!t]
\centering
\caption{\textbf{Spatial response} by point group
(a.u., mean $\pm$ SD over presses).}
\label{tab:spatial_response}
\footnotesize
\setlength{\tabcolsep}{3pt}
\begin{tabular}{@{}lcccc@{}}
\toprule
 & Group 1 & Group 2 & Group 3 & Group 4 \\
 & ($n{=}20$) & ($n{=}40$) & ($n{=}40$) & ($n{=}40$) \\
\midrule
Normal $z$
 & $646.1{\pm}68.7$
 & $992.7{\pm}79.5$
 & $588.8{\pm}82.0$
 & $683.6{\pm}79.0$ \\
Shear $(x,y)$
 & $92.3{\pm}3.4$
 & $146.3{\pm}17.6$
 & $91.4{\pm}8.7$
 & $59.4{\pm}10.6$ \\
\bottomrule
\end{tabular}
\end{table} 

\section{Learning Touch-centric Manipulation}
\label{sec:learning}

Our experiments address two questions: whether dense three-axis tactile feedback improves manipulation performance, and how spatial contact information and shear feedback contribute to different manipulation behaviors.
We compare \textbf{PanoTac} with a vision-only baseline and visuo-tactile baselines using different tactile inputs, analyzing full-task success, stage-level performance, and failure cases.

\begin{figure*}[t]
    \centering
    \includegraphics[width=\linewidth]{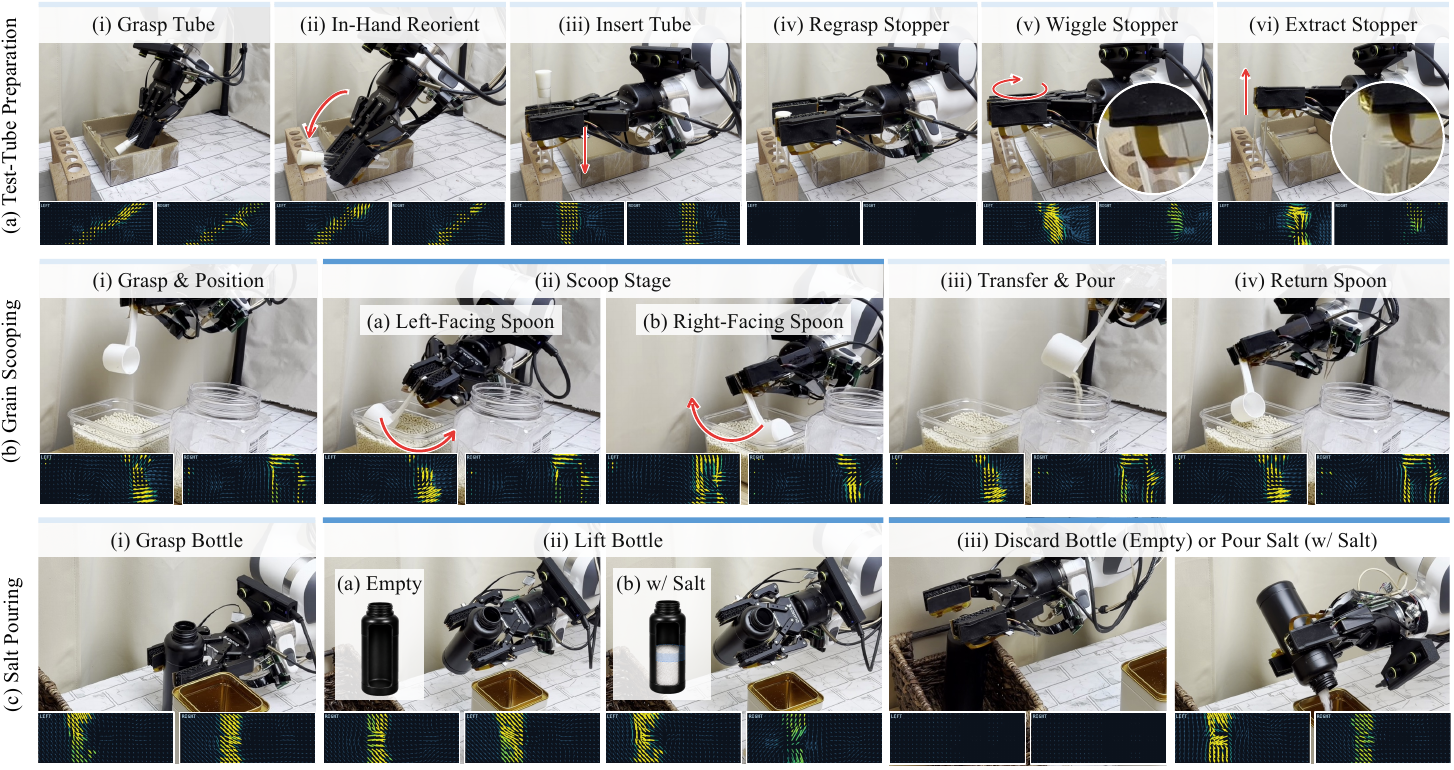}
    \caption{\textbf{Qualitative Results.}
Representative stages of (a) test-tube preparation,
(b) grain scooping, and (c) salt pouring.
Paired tactile visualizations show responses from both fingertips,
and red arrows indicate motion directions.
Blue bars mark stages shown for both task branches.
Detailed task descriptions are provided in Sec.~\ref{sec:learning}.}
    \label{fig:task_progress}
\end{figure*}


\begin{table*}[t]
  \centering
  \caption{\textbf{Quantitative Results.}
    Success rates (\%) over 20 rollouts per policy per task,
    with 10 per spoon orientation or bottle condition.
    Overall reports full-task success averaged across the
    three tasks. Stage rates are not conditioned on reaching
    the stage. Bold indicates the best result in each column.}
  \label{tab:modality_ablation}

  \begingroup
  \footnotesize
  \setlength{\tabcolsep}{2pt}
  \renewcommand{\arraystretch}{1.25}
  \setlength{\aboverulesep}{0.4ex}
  \setlength{\belowrulesep}{0.4ex}

  \newcommand{\tblhead}[1]{%
    {\scriptsize\shortstack[c]{#1}}%
  }

  \begin{tabularx}{\textwidth}{
    @{}
    >{\raggedright\arraybackslash}p{0.23\textwidth}
    *{11}{>{\centering\arraybackslash}X}
    @{}
  }
    \toprule
    \multirow[c]{3}{*}{\textbf{Policy}}
      & \multicolumn{3}{c}{
          \tblhead{\textbf{Test-Tube Preparation}}}
      & \multicolumn{4}{c}{\textbf{Grain Scooping}}
      & \multicolumn{3}{c}{\textbf{Salt Pouring}}
      & \multirow[c]{3}{*}{\tblhead{\textbf{Overall}}} \\
    \cmidrule(lr){2-4}
    \cmidrule(lr){5-8}
    \cmidrule(lr){9-11}
      & \multirow[c]{2}{*}{\tblhead{Reorient}}
      & \multirow[c]{2}{*}{\tblhead{Insert}}
      & \multirow[c]{2}{*}{
          \tblhead{\textbf{Unplug}\\\textbf{(END)}}}
      & \multicolumn{2}{c}{\tblhead{Pre-Scoop Direction}}
      & \multirow[c]{2}{*}{\tblhead{Scoop}}
      & \multirow[c]{2}{*}{
          \tblhead{\textbf{Return}\\\textbf{(END)}}}
      & \multicolumn{2}{c}{\tblhead{Sort Bottle Stage}}
      & \multirow[c]{2}{*}{
          \tblhead{\textbf{Return}\\\textbf{(END)}}}
      & \\
    \cmidrule(lr){5-6}
    \cmidrule(lr){9-10}
      & & &
      & \tblhead{Left}
      & \tblhead{Right}
      & &
      & \tblhead{With Salt}
      & \tblhead{Empty}
      & & \\
    \midrule

    Vision Only
      & \textbf{100} & 90 & 35
      & 20 & 80 & 50 & 45
      & 50 & 60 & 55
      & 45.0 \\

    Vision w/ Dense Normal Array
      & \textbf{100} & \textbf{100} & 65
      & 80 & \textbf{100}
      & \textbf{85} & \textbf{85}
      & 90 & 50 & 70
      & 73.3 \\

    Vision w/ Sparse Three-Axis
      & \textbf{100} & \textbf{100} & 85
      & 90 & 70 & 70 & 60
      & 90 & \textbf{100} & 90
      & 78.3 \\

    Vision w/ PanoTac
      & \textbf{100} & \textbf{100} & \textbf{90}
      & \textbf{100} & \textbf{100}
      & \textbf{85} & \textbf{85}
      & \textbf{100} & \textbf{100} & \textbf{100}
      & \textbf{91.7} \\

    \bottomrule
  \end{tabularx}
  \endgroup
\end{table*}

\subsection{Experimental Setup}
\label{ssec:learning_setup}

We evaluate all policies on three touch-centric manipulation tasks:
\textit{(i) test-tube preparation}, a long-horizon task in which dense contact patterns support tube reorientation and insertion, while shear feedback helps detect stopper release;
\textit{(ii) grain scooping}, in which contact patterns along the spoon handle help infer spoon orientation and guide the scooping direction;
and \textit{(iii) salt pouring}, in which shear feedback provides weight cues from an opaque bottle to guide the choice between pouring and discarding.

For each task, we train four policies on the same demonstrations: vision only, vision with dense piezoresistive contact maps, vision with sparse three-axis magnetic readings, and vision with \textit{PanoTac}'s dense three-axis tactile representation. We refer to these as the vision-only, dense normal, sparse three-axis, and PanoTac policies, respectively. Each policy is evaluated over 20 rollouts per task, with at most 800 steps per rollout. Stage-level success rates are computed over all rollouts for the corresponding task or condition, including those that fail before reaching the stage. Overall success is the full-task success rate pooled across all three tasks (60 rollouts per policy). Task details follow.

\subsection{Test-Tube Preparation}

\subsubsection{Task Setup} 
The robot retrieves a test tube from a storage box, reorients
it upright against a rack, and inserts it into the rack. It then wiggles the stopper to loosen it, extracts it from the tube, and returns it to the box. Contact patterns provide cues about the tube's
in-hand orientation, while shear feedback helps detect
stopper release. We collect 50 demonstrations varying tube location and height, stopper color, and stopper fit tightness. Variations in fit tightness change the force required for removal. Fig.~\ref{fig:task_progress}(a) shows a
representative rollout. \textbf{Evaluation Metrics:}
Reorientation succeeds when the tube is upright.
Insertion succeeds when the tube is placed in the rack.
Unplugging succeeds when the stopper is fully removed from the tube before the robot begins the return motion.
A premature return attempt is counted as an unplugging failure.
Every policy returns the stopper to the box in all rollouts that reach this stage, so we report unplugging as the final stage.

\subsubsection{Analysis and Results}
This task examines whether tactile feedback supports tube alignment and the transition from stopper loosening to extraction.

\textbf{Tactile-refined alignment.}
All four policies achieve $100\%$ reorientation success.
Insertion additionally requires aligning the tube with the rack opening.
All three tactile policies achieve $100\%$ insertion success, compared with $90\%$ for the vision-only policy (Table~\ref{tab:modality_ablation}), consistent with tactile feedback supporting alignment during insertion.

\begin{figure*}
    \centering
    \includegraphics[width=\linewidth]{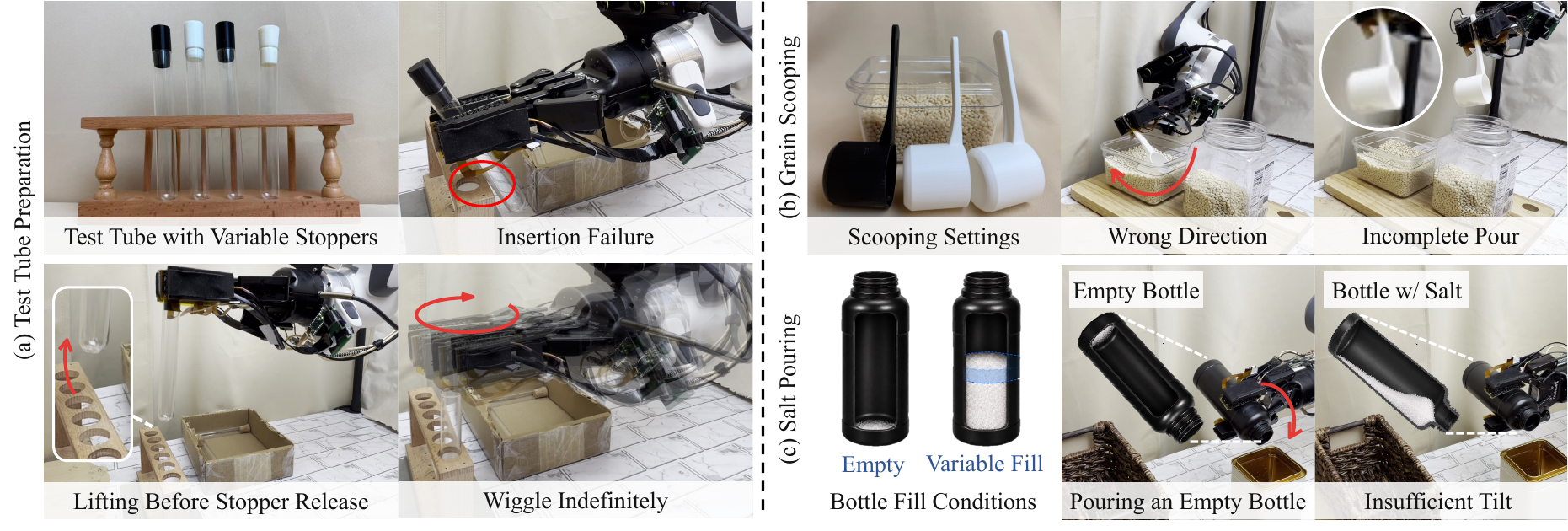}
    \caption{\textbf{Representative Failure Cases of Baseline Policies.} We show variations in (a) test tubes, (b) spoons, and (c) bottle fill levels, together with representative manipulation failures. Detailed failure analysis is provided in Sec.~\ref{sec:learning}.}
    \label{fig:failure_cases}
    \vspace{-1.5em}
\end{figure*}

\textbf{Force-sensitive wiggling.}
This stage requires applying sufficient force to loosen the stopper and detecting when to transition from wiggling to extraction.
PanoTac achieves $90\%$ unplugging success, compared with $35\%$ for the vision-only policy, $65\%$ for the dense normal policy, and $85\%$ for the sparse three-axis policy.
Representative vision-only failures involve lifting
before stopper release, which can pull the tube out of
the rack or cause it to slip from the grasp
(Fig.~\ref{fig:failure_cases}(a)).
The dense normal policy is the only policy observed to continue wiggling after stopper release (Fig.~\ref{fig:failure_cases}(a)), revealing difficulty detecting stopper release and coordinating the transition to extraction using dense normal feedback alone.
Both policies with magnetic feedback achieve higher unplugging success than the dense normal policy, consistent with magnetic responses providing useful cues for stopper release.
PanoTac achieves the highest observed unplugging
success, with one more successful rollout than the
sparse three-axis policy.

\subsection{Grain Scooping}

\subsubsection{Task Setup}
The robot must scoop grain from one container, pour it into
another, and return the spoon. The spoon initially faces
either left or right, requiring different scooping directions.
After grasping, the spoon bowl is \textbf{occluded from the wrist
camera}, while contact patterns along the handle provide
orientation cues. We collect 45 demonstrations varying spoon color, handle
length and width, spoon placement, grain surface level,
and container height. The container is raised by
$3$\,cm in approximately half of the demonstrations,
and the two spoon orientations are balanced. These variations
require the robot to adjust its motion to reach the grain.
Fig.~\ref{fig:task_progress}(b) shows a representative rollout.
\textbf{Evaluation Metrics:}
Direction selection succeeds when the robot initiates
scooping in the direction appropriate for the spoon's
orientation.
Scooping succeeds when grain is lifted from the source
container.
Return succeeds when the spoon
is returned to the source container.

\subsubsection{Analysis and Results}
This task examines how dense contact geometry guides the scooping
direction and how tactile feedback supports execution under
varying grain levels and container heights. 

\textbf{Tool-geometry-based action selection.}
The policy must infer the spoon's orientation from its
in-hand contact pattern. Moving in the wrong
direction is the dominant failure of the
baseline policies (Fig.~\ref{fig:failure_cases}(b)).
Averaged over both spoon orientations, direction
selection succeeds in $100\%$ of rollouts for PanoTac,
$90\%$ for the dense normal array policy, $80\%$ for the
sparse three-axis policy, and $50\%$ for vision only
(Table~\ref{tab:modality_ablation}).
The dense normal array policy selects the wrong
direction in only two of 20 trials, suggesting that dense
contact patterns are informative for handle orientation.
Magnetic responses may also provide orientation cues,
since the sparse three-axis policy performs above chance.
PanoTac combines both and selects the correct direction
in all 20 rollouts. The two modalities appear to resolve
ambiguities that remain when either is used alone.
The vision-only policy performs at chance level,
with a bias toward scooping to the right.

\textbf{Contact-aware execution.}
Full-task success favors the policies with dense
contact information: PanoTac and the dense normal
array policy both achieve $85\%$, compared with
$60\%$ for the sparse three-axis policy and $45\%$ for
vision only.
Choosing the right direction does not guarantee successful
grain transfer. Even when the robot moves spoon in the intended
direction, it may pass through the grain without collecting
a full scoop, and pouring may leave residual grain in the spoon.
The sparse three-axis policy also loses track of the
task stage. In one rollout it pours twice, although
every demonstration contains a single pour. In another,
it begins returning the spoon before the grain is fully
dispensed (Fig.~\ref{fig:failure_cases}(b)).
Its sparse magnetic feedback may not always reveal which
stage the task is in.
Return failures occur in both this policy and the
vision-only policy.
For the vision-only policy, one possible explanation is
the variation in container height across demonstrations. An action suited to a lower container
leaves the spoon poorly positioned for return, and the
camera sees too little below the gripper to correct it.


\subsection{Salt Pouring}

\subsubsection{Task Setup}
The opaque bottle's \textbf{contents are fully occluded from view}.
The robot must lift the bottle and use its weight
to choose between pouring and discarding. If the bottle
contains salt, the robot pours it into an opaque central
container and returns the bottle to its original position.
Otherwise, it discards the bottle into a trash basket.
Fig.~\ref{fig:task_progress}(c) shows a representative rollout.
We collect 40 demonstrations: 20 with near-empty bottles
containing residual salt and 20 with variable amounts of
salt, approximately half a bottle. Half a bottle of salt
weighs around 150 grams. Each pouring episode
is followed by an empty-bottle episode that retains any
leftover salt. Evaluation follows a scrambled order, with
10 rollouts per bottle condition. \textbf{Evaluation Metrics:}
We evaluate bottle sorting and final placement.
For an empty bottle, sorting succeeds if the bottle is
tilted no further than the onset of pouring. For a bottle
with salt, sorting succeeds if the robot produces a steady
stream and begins returning the bottle only after the
stream stops. Full-task success additionally requires
returning the bottle to its original position after
pouring, or discarding it into the basket when empty.

\subsubsection{Analysis and Results}
This task examines whether tactile shear cues support
the choice between pouring and discarding when the
bottle's contents are visually hidden.

\textbf{Weight-based action selection.}
The PanoTac policy achieves $100\%$ sorting success,
compared with $95\%$ for the sparse three-axis policy,
$70\%$ for the dense normal array policy, and $55\%$
for vision only (Table~\ref{tab:modality_ablation}).
The strong performance of both policies with magnetic
feedback supports its utility for distinguishing bottle
conditions through their shear responses.
The sparse three-axis policy fails once in sorting and
once in placement, likely because its shear response
varies with grasp location. PanoTac's dense contact map
localizes the grasp, which helps disambiguate these
readings.
The dense normal array policy sorts $90\%$ of bottles
with salt but only $50\%$ of empty bottles, attempting
to pour in every empty-bottle failure
(Fig.~\ref{fig:failure_cases}(c)).
This bias may reflect the greater length of pouring
trajectories.
Its single failure with salt is an insufficient tilt
that releases no salt (Fig.~\ref{fig:failure_cases}(c)),
possibly because a weak weight signal blurs the boundary
between pouring and discarding, leading the policy to
average the two actions.

\textbf{Gripper compliance may provide indirect weight cues.}
In informal trials with a bottle filled to capacity, we observe
successful vision-only execution, suggesting that larger
weight-induced deformation may provide a visible weight cue.
Smaller deformations at the evaluated salt levels could
likewise affect the visual observations and normal contact
map, although these cues are difficult to discern by eye.
This offers a possible explanation for some successful
decisions without magnetic feedback.
Appendix~\ref{app:bonus} extends this task to pouring liquids, where we test out-of-distribution capability of our policy by including liquids of unseen color and couscous, which have a different pouring dynamics.

\section{Conclusion}
In this paper, we present \textbf{PanoTac}, a low-cost, compliant tactile fingertip for dense three-axis tactile sensing. Hardware characterization showed spatially consistent and temporally repeatable responses across loads and sensor instances. Integrated into a visuo-tactile learning pipeline, our policies outperformed vision-only baselines by an average of 46.7 percentage points on three manipulation tasks, and modality ablations indicated that spatial contact information and shear feedback each contribute to performance. Given its cost, fabrication simplicity, and customizable form factor, we believe \textbf{PanoTac} is a practical tactile interface for manipulation.

\textbf{Limitations.} First, the magnetometer readings are susceptible to ferromagnetic objects and external magnetic fields; training on interaction data that include such objects is the most direct remedy. Second, force resolution is bounded by the compliance of the elastomer and the number of magnetometers; scaling the magnetometer array and tuning the elastomer are natural next steps. Third, the current fingertip is shaped for a parallel-jaw gripper; adapting it to dexterous hands and evaluating in-hand manipulation would test the representation beyond two-finger grasping. Finally, the sensor's low cost and simple installation make it a candidate for handheld, UMI-style data collection, which we leave to future work.

\section*{ACKNOWLEDGMENT}
%

We thank Venkatesh Pattabiraman, Zifan Zhao, Bowen Tan, Hengkai Pan, Kanad Patel,
Yibo Yan, Nikhil Bhattasali, and Kevin Wu for helpful discussions.

We used ChatGPT (OpenAI) and Claude Fable (Anthropic) to assist with language polishing and Blender to create the 3D illustration in Fig.~\ref{fig:hardware}. 


\bibliographystyle{IEEEtran}
\bibliography{references}


\clearpage
\let\appendixsection\section
\useRomanappendicesfalse
\appendices
\let\section\appendixsection
\renewcommand{\thesectiondis}{\thesection.}
\renewcommand{\thesubsection}{\thesection.\arabic{subsection}}
\renewcommand{\thesubsectiondis}{\thesubsection}
\twocolumn[{%
  \centering
  {\LARGE\bfseries Appendix\par}
  \vspace{1.5em}
}]

\section{Teleoperation Interface}
\label{app:teleop}

The demonstrations are collected through VR teleoperation with a Meta Quest~3
at 30\,Hz.

\textit{Tactile feedback.} To help the operator regulate contact during
manipulation, we present the tactile measurements as visual and auditory
feedback. A real-time heatmap of each fingertip's contact map shows the
spatial distribution of contact. In parallel, the left and right
fingertips drive tones in the corresponding stereo channels, allowing
the operator to distinguish force changes at the two fingers.
For each fingertip, we compute the Euclidean norm of the deviation of
its 30 magnetometer values (10 three-axis readings) from their
exponential moving average. Values below a deadzone produce silence;
above it, the signal is logarithmically compressed and mapped to
increasing volume and pitch over 300--1200\,Hz. The audio therefore
emphasizes changes in force rather than its absolute level.

\textit{Recording.} For each demonstration, we record wrist-camera RGB
images, robot proprioception, tactile measurements from both fingertips,
and the commanded end-effector poses and gripper actions. Tactile
measurements are acquired at 60\,Hz, and the recorded streams are aligned
at 30\,Hz. We process the tactile measurements into the three-axis
representation described in Sec.~\ref{subsec:signal}; policy inputs and
subsampling are detailed in Appendix~\ref{app:policy}.

\section{Policy Implementation}
\label{app:policy}

All four policies in Sec.~\ref{sec:learning} share the same vision encoder,
diffusion backbone, and training recipe; they differ in their tactile
inputs and encoders, which are absent in the vision-only policy.

\subsection{Observations and Encoders}
\label{app:obs}

Table~\ref{tab:app_obs} lists the per-frame inputs and their encoders.
The policy conditions on a single observation frame.

\begin{table}[h]
\centering
\caption{\textbf{Observation inputs and encoders.} Shapes are per frame;
L and R denote the two fingertips.}
\label{tab:app_obs}
\footnotesize
\setlength{\tabcolsep}{3pt}
\begin{tabular}{@{}llll@{}}
\toprule
\textbf{Modality} & \textbf{Input} & \textbf{Encoder} & \textbf{Dim} \\
\midrule
RGB image       & $3 \times 256 \times 256$      & ResNet-18             & 512 \\
Proprioception  & $9$ (position, 6D rotation)   & identity              & 9 \\
\addlinespace[2pt]
Dense normal    & $2 \times 12 \times 32$        & CNN                   & 128 \\
Sparse 3-axis   & $2 \times 10 \times 3$         & MLP                   & 256 \\
PanoTac         & $3 \times 24 \times 32$        & CNN                   & 512 \\
\bottomrule
\end{tabular}
\end{table}

\textit{Vision.} The image encoder is an ImageNet-pretrained ResNet-18,
fine-tuned jointly with the policy.

\textit{PanoTac representation.} The fields $\mathbf{T}$ of the two
fingertips (Sec.~\ref{subsec:signal}) are stacked along the row axis into
a single $3 \times 24 \times 32$ image.

\textit{Conditioning.} The vision, tactile, and proprioceptive features
are concatenated into a single vector, which conditions every residual
block of the U-Net through FiLM~\cite{film}.

\subsection{Diffusion Policy and Training}
\label{app:training}

We use the convolutional U-Net diffusion policy of~\cite{diffusionpolicy}
with the hyperparameters in Table~\ref{tab:app_hparams}.

\begin{table}[h]
\centering
\caption{\textbf{Diffusion policy hyperparameters}, shared across all
policies and tasks.}
\label{tab:app_hparams}
\footnotesize
\setlength{\tabcolsep}{4pt}
\begin{tabular}{@{}ll@{}}
\toprule
\textbf{Parameter} & \textbf{Value} \\
\midrule
U-Net channels           & $[512, 1024, 2048]$ \\
Diffusion steps          & 100 (train), 16 DDIM (inference) \\
Noise schedule           & squared cosine, $\epsilon$-prediction \\
Observation horizon      & 1 frame \\
Prediction horizon       & 16 steps ($\approx 2.7$\,s at 6\,Hz) \\
Optimizer                & AdamW, lr $10^{-4}$, weight decay $0.01$ \\
LR schedule              & 10-epoch warmup, cosine decay \\
Batch size / epochs      & 128 / 150 \\
EMA                      & on U-Net, max decay 0.9999 \\
\bottomrule
\end{tabular}
\end{table}

\subsection{Actions and Deployment}
\label{app:deploy}

The policy predicts a 10-dimensional absolute action: end-effector
position (3), rotation in the 6D representation (6), and a binary gripper
command (1), normalized to $[-1, 1]$. Recorded streams are aligned at
30\,Hz and subsampled by a factor of 5, so the policy operates at 6\,Hz.
At deployment, the policy runs inference at every step, and overlapping action
chunks are combined by exponential temporal ensembling with a half-life
of 4 policy steps, weighting the most recent prediction highest. The
ensembled action is interpolated to 30\,Hz for the robot controller.

\section{Sensor Characterization Details}
\label{app:hardware}

This section details the procedures behind Sec.~\ref{sec:characterization}.
In both experiments, the fingertip lies flat on the bench surface.

\subsection{Response Computation}
For each recording, we average both sensing streams over the recording
window, subtract the no-contact baseline, and compute $\mathbf{T}$ as in
Sec.~\ref{subsec:signal}. The normal response is the Frobenius norm
$\lVert N \rVert_F$ of the normal map in~\eqref{eq:normal_map}. The shear
response is the RMS of $\lVert (S^x_{ij}, S^y_{ij}) \rVert$ over all
$12 \times 32$ taxels, where $S^x$ and $S^y$ are obtained by
inverse-distance-weighted projection of the magnetometer readings, as in Sec.~\ref{subsec:signal}.

\subsection{Load Response}
Each instance undergoes three sweeps over 15 loads in increasing
order: 0.4, 0.6, 0.8, 1.0, 1.2, 1.5, 2.0, 2.5, 3.0, 3.5, 4.0, 6.0, 8.0,
10.0, and 12.0\,N. Before each sweep, with the indenter lifted, we record a one-second no-contact baseline. At each load, the indenter is driven past the target force and backed off until the gauge reads the target. We hold the load until the readings stop drifting, then record both layers for one second. The indenter stays fixed within each sweep.

\subsection{Spatial Response}
The spherical indenter is pressed by hand onto the fingertip. Before every press, with the indenter resting just above the site, we record a two-second no-contact baseline, which removes drift between presses. The site is then loaded to 5\,N and held while both layers are recorded for two seconds. Each site is pressed 10 times, and the resulting presses are pooled within each group listed in Table~\ref{tab:spatial_response}.

\section{Bonus Task: Pour Anything}
\label{app:bonus}
The salt-pouring task shows that PanoTac uses shear cues to distinguish filled from empty bottles and decide whether to pour, much as humans judge a bottle’s contents by lifting it without seeing inside. We next ask whether this capability extends to different visual conditions and contents. As a bonus task, we consider liquid pouring, where spills have greater consequences. We train a policy using 36 demonstrations and evaluate it on unseen liquid colors and unseen couscous, testing whether the learned pour-or-discard behavior generalizes across appearance, density, and flow.
\begin{figure}[t]
\centering
\includegraphics[width=\linewidth]{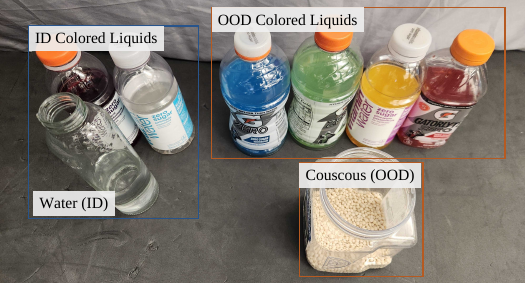}
\caption{\textbf{Tested Contents.} In-distribution (ID) liquids comprise water and colored liquids; out-of-distribution (OOD) contents comprise colored liquids and couscous. The retail bottles and jar display the contents;
the robot uses the small transparent bottle shown in front.}
\label{fig:bonus_contents}
\end{figure}

As in salt pouring, the robot must lift a bottle and choose between pouring and discarding. If the bottle contains material, the robot pours it into the receiving beaker and returns the bottle to its starting position; otherwise, it discards the bottle into the basket.
Although the bottle is transparent, \textbf{the wrist camera cannot see the liquid until after the policy decides to pour}. The pour-or-discard decision therefore has the same visual occlusion as in the salt task.
Fig.~\ref{fig:bonus_contents} shows the liquids and couscous used in these trials.

\begin{figure}[t]
\centering
\includegraphics[width=\linewidth]{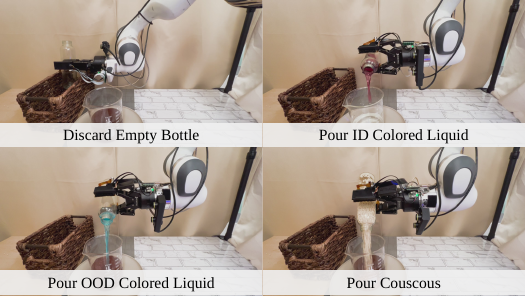}
\caption{\textbf{Bonus Task Rollouts.} Representative frames show
empty-bottle discard, ID and OOD colored-liquid pouring, and couscous
pouring.}
\vspace{-2em}
\label{fig:bonus_execution}
\end{figure}

We report all 15 consecutive autonomous trials from a single
continuous 9\,min\,44\,s recording, which also captures manual preparation
and resets between trials. The website presents these trials in a
$2\times2$ grid, grouped as empty, ID, OOD color, and OOD material. Fig.~\ref{fig:bonus_execution} shows
representative frames. We show only one couscous trial, at the end,
because the grains cannot be cleaned out for repeated trials, as shown in the continuous recording.

Success requires discarding an empty bottle without pouring, or pouring
its contents into the beaker and returning the bottle.
Table~\ref{tab:bonus_results} reports successful/recorded trials, pooling OOD liquids and couscous. All 15 trials succeed.

\begin{table}[htbp]
\centering
\caption{\textbf{Bonus Task Results.} Successful/recorded trials by condition
and overall success rate in the continuous recording.
ID comprises two water trials, one dark-red-liquid trial, and one
faint-white-liquid trial. OOD pools four colored-liquid trials and one
couscous trial.}
\label{tab:bonus_results}
\small
\setlength{\tabcolsep}{6pt}
\begin{tabular}{@{}lcccc@{}}
\toprule
\textbf{Policy} & \textbf{Empty} & \textbf{ID} & \textbf{OOD} & \textbf{Total} \\
\midrule
PanoTac & 6/6 & 4/4 & 5/5 & 100\% \\
\bottomrule
\end{tabular}
\end{table}

\end{document}